\documentclass[10pt,twocolumn,letterpaper]{article}

\usepackage[pagenumbers]{paper}    

\usepackage{iftex}
\ifXeTeX
  \usepackage[T1]{fontenc}
\fi
\usepackage{graphicx}
\usepackage{booktabs}
\usepackage{amsmath}
\usepackage{amssymb}
\usepackage{multirow}
\usepackage{tikz}
\usetikzlibrary{calc,positioning}
\usepackage{standalone} 

\usepackage{capt-of} 

\definecolor{linkblue}{rgb}{0.21,0.49,0.74}
\usepackage[breaklinks,colorlinks,allcolors=linkblue]{hyperref}

\title{InfScene-SR: Seamless Super-Resolution of Arbitrarily Large Remote-Sensing Scenes via Variance-Preserving Joint Denoising}

\author{Shoukun Sun\\
University of Idaho\\
Moscow, Idaho, USA\\
{\tt\small ssun@uidaho.edu}
\and
Zhe Wang\\
NCEAS, UC Santa Barbara\\
Santa Barbara, California, USA\\
{\tt\small zwang@nceas.ucsb.edu}
\and
Xiang Que\\
Fujian Agriculture and Forestry University\\
Fuzhou, China\\
{\tt\small quexiang@fafu.edu.cn}
\and
Jiyin Zhang\\
University of Idaho\\
Moscow, Idaho, USA\\
{\tt\small jiyinz@uidaho.edu}
\and
Xiaogang Ma\\
University of Idaho\\
Moscow, Idaho, USA\\
{\tt\small max@uidaho.edu}
}

\begin{document}
\twocolumn[{%
\renewcommand\twocolumn[1][]{#1}%
\maketitle
\begin{center}
    \includegraphics[width=\textwidth]{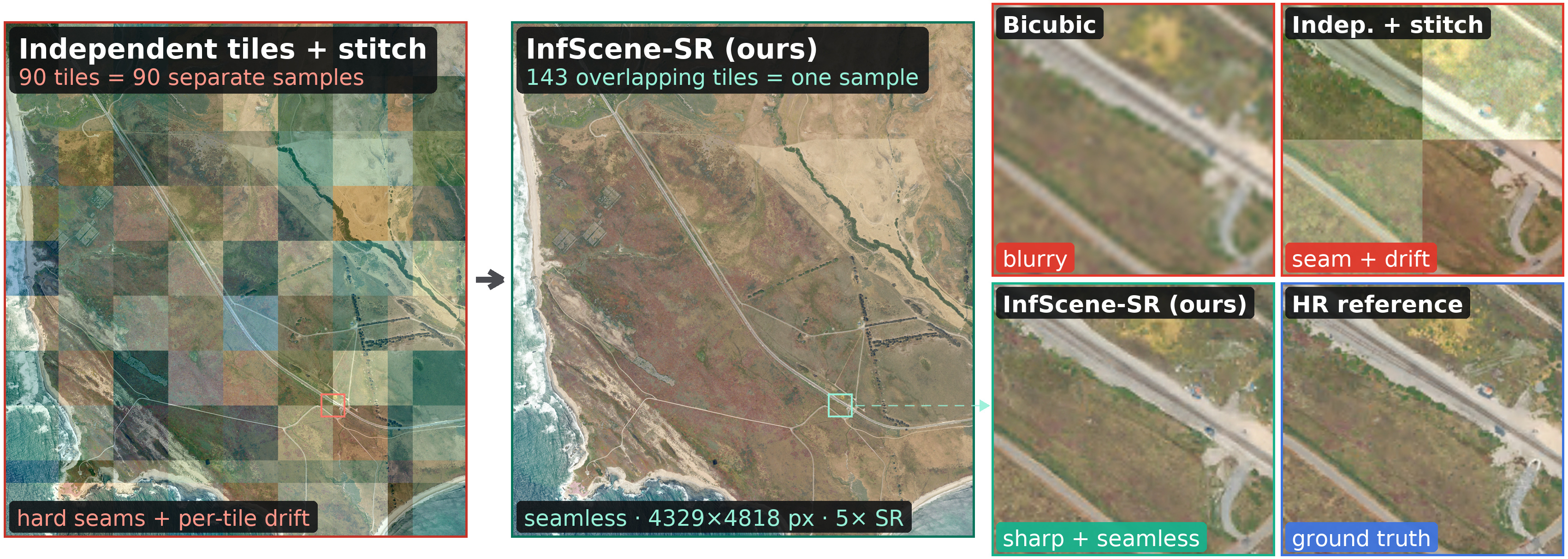}
    \captionof{figure}{5$\times$ SR of the 21-megapixel NAIP test scene. Stitching independently denoised tiles leaves seams and per-tile drift (left); InfScene-SR denoises 143 overlapping tiles jointly into one seamless, sharp sample (center). Right: detail at a four-tile corner---bicubic is blurry, stitching breaks, ours is sharp and continuous, matching native HR.}
    \label{fig:teaser}
\end{center}%
}]
\begin{abstract}
   Diffusion models now give the best perceptual quality in super-resolution (SR), but their architecture and training confine them to small fixed crops. Operational remote sensing needs seamless scenes orders of magnitude larger. Joint denoising fuses overlapping tiles at every reverse step and lets text-to-image diffusion generate beyond its training crop, but it assumes deterministic ODE samplers. With the stochastic sampler of SR models such as SR3, the averaging also partly cancels independent per-tile noise. This known variance erosion blurs the detail SR should recover and must be corrected. We carry variance-corrected fusion to conditional SR and derive \emph{Spatially-Decoupled Variance Correction} (SDVC), an exact reformulation that replaces per-step global normalization with independent per-tile contributions accumulated in one additive pass. SDVC turns the coupled per-step computation into independent tile-local work, so the resulting pipeline, InfScene-SR, runs in parallel across GPUs and makes SR of arbitrarily large scenes feasible. On a 5$\times$ SR task built from NAIP aerial imagery, we evaluate whole scenes with fidelity, perceptual, seam-continuity, and faithfulness metrics. Under one backbone, InfScene-SR is the only fusion strategy that is seamless and sharp at once, the closest to the low-resolution observation among those that synthesize detail, and within 0.003 IoU of native high-resolution imagery on downstream invasive-plant segmentation. Code is available at \url{https://github.com/TitorX/infscene-sr}.
\end{abstract}

\section{Introduction}
\label{sec:intro}

Single-image super-resolution (SR) recovers a high-resolution image from one low-resolution observation. The problem is ill-posed, because many high-resolution images reduce to the same low-resolution input, so an SR model has to synthesize the missing detail while staying consistent with the observation. Remote sensing is a natural application. Spatial resolution trades against coverage and revisit frequency, and SR offers a way to lift routinely collected coarse imagery toward the detail of a sharper but rarer sensor. The imagery is also large. An aerial survey of one county spans billions of pixels, and whole-slide pathology or microscopy mosaics are similar. Geospatial workflows such as mosaicking, object delineation, and change detection operate on the continuous raster, so tile-boundary artifacts propagate directly into scientific measurements. The output that matters is a \emph{seamless scene}.

SR models come in two kinds. Feed-forward networks, from CNN regression~\cite{srcnn_2016} and GANs~\cite{srgan_2017,esrgan_2018} to transformers~\cite{swinir_2021,hat_2023}, map an image of any size to its output in one deterministic pass, so they super-resolve large images by simple tiling. Diffusion models such as SR3~\cite{sr3_2023} instead generate the output by iterative denoising on a fixed crop, drawing one sample from the distribution of high-resolution images consistent with the input. We choose diffusion for the quality of that sample. A network trained with a pixel loss returns the average of all plausible reconstructions and blurs, whereas a diffusion model samples one of them and stays sharp. In human studies~\cite{sr3_2023,stablesr_2024} and perceptual metrics~\cite{seesr_2024}, on natural and remote-sensing imagery~\cite{ediffsr_2024}, diffusion models lead regression and GAN networks in perceptual quality. Their practical limitation is scale. The architecture fixes the working size, and training on whole scenes is infeasible, so diffusion SR works on small crops (e.g., $512{\times}512$), a tiny fraction of a survey scene. We therefore ask how a fixed-crop diffusion model can super-resolve a whole scene, rather than settling for a feed-forward network.

The obvious workaround (partition, super-resolve tiles independently, stitch) fails for diffusion SR. The reverse process is stochastic, so adjacent tiles are \emph{independent samples} whose content, texture, and radiometry need not agree, producing hard seams and tile-to-tile offsets in brightness and color (Fig.~\ref{fig:teaser}). Feathered blending of finished tiles softens edges but cannot reconcile content that has already diverged.

A principled alternative from large-content text-to-image generation is \emph{joint denoising}~\cite{multidiffusion_2023,syncdiffusion_2023}: overlapping tiles are denoised in lockstep, with overlaps fused at every reverse step. These methods, however, assume deterministic ODE-type samplers (e.g., DDIM~\cite{ddim_2020}), whereas SR3 uses \emph{stochastic} ancestral sampling. Averaging overlaps then erodes the sampling variance. Independent noise realizations partially cancel, leaving the fused state with a deficit that is re-introduced at every step. Sun~\etal~\cite{gvcf_2025} identified this deficit in text-to-image generation and proposed variance-corrected fusion (VCF) to repair it. We carry the analysis and the correction to conditional SR, where the LR condition anchors every tile.

Correcting the variance does not yet make gigapixel SR \emph{practical}. The corrected fusion couples every pixel to all covering tiles through global normalization, so a direct implementation gathers all overlapping predictions into one memory space at each of $T$ steps. Our key observation is that this coupling can be removed without approximation: the fusion can be rewritten so that each tile's update is a self-contained local contribution and the global state is maintained by pure accumulation. We call this \emph{Spatially-Decoupled Variance Correction} (SDVC). Although SDVC is an identity, it changes the systems picture: constant per-device memory, a resumable disk-streamed canvas, and asynchronous multi-worker execution.

We ground the method in an operational gap. NAIP aerial imagery offers 0.6\,m resolution every 2--3 years, PlanetScope ${\sim}3$\,m daily. Lifting 3\,m imagery to sub-meter quality would enable high-frequency monitoring at NAIP-like fidelity (Sec.~\ref{sec:datasets}). We train a 5$\times$ SR3 model on 236k NAIP patches and evaluate \emph{whole scenes}. Alongside fidelity and perceptual metrics, we introduce a seam-continuity metric defined on the inference tiling, adopt an LR-consistency (faithfulness) metric, and test whether super-resolved scenes support a real downstream task, segmentation of the invasive iceplant (\textit{Carpobrotus edulis}).

Our contributions are:
\begin{itemize}
    \item \textbf{Analysis.} We extend the variance-erosion analysis of~\cite{gvcf_2025} to conditional SR with stochastic samplers: it explains why ODE-style averaging fails for SDE-based SR (Sec.~\ref{sec:method}).
    \item \textbf{System.} We derive SDVC, an exact spatial decoupling of the corrected fusion, and show its consequences: constant per-device memory, resumable disk-streamed state, and asynchronous multi-GPU execution (Secs.~\ref{sec:sdvc}--\ref{sec:cost}, \ref{sec:efficiency}).
    \item \textbf{Evaluation.} A whole-scene protocol for tiled SR---seam-continuity and faithfulness metrics plus a controlled comparison where \emph{only} the fusion rule varies---on which InfScene-SR is the only strategy that is seamless and sharp at once, departs least from the LR observation among strategies that synthesize detail, and nearly matches native HR imagery on downstream iceplant segmentation (Sec.~\ref{sec:results}).
\end{itemize}

\section{Related Work}
\label{sec:related}

\begin{figure*}[!t]
    \centering
    \resizebox{\textwidth}{!}{\input{figs/pipeline_sdvc.tex}}
    \caption{Reverse step of InfScene-SR. The scene canvas $y_t$ is cropped into overlapping views; per-device workers run one sampler step and emit a tile-local SDVC contribution $C_t^{(i)}$ (Eq.~\ref{eq:sdvc}), accumulated additively, into the canvas for step $t{-}1$. Repeated for $t=T,\dots,1$, the scene evolves as one consistent sample; independent per-tile denoising with stitching (bottom right) leaves seams and per-tile drift.}
    \label{fig:pipeline}
\end{figure*}

\paragraph{Single-image super-resolution.}
Deep SISR progressed from direct regression (SRCNN~\cite{srcnn_2016}) through residual and dense architectures~\cite{resnet_2016,densenet_2017} to transformer designs~\cite{attention_2017} such as SwinIR~\cite{swinir_2021} and HAT~\cite{hat_2023}. All of these are feed-forward: the output is a deterministic local function of the input, so any image size can be processed, or tiled and stitched without visible seams. Pixel-wise losses reward the conditional mean and look blurry; adversarial training (SRGAN~\cite{srgan_2017}, ESRGAN~\cite{esrgan_2018}) sharpens at the cost of stability. This tension is formalized as the perception--distortion trade-off~\cite{blau_perception-distortion_2018}, central to interpreting our results (Sec.~\ref{sec:pd}).

\paragraph{Diffusion models for super-resolution.}
DDPMs~\cite{ho_denoising_2020} learn to invert a gradual noising process. SR3~\cite{sr3_2023} conditions that reverse process on the LR image and produces reconstructions that human raters prefer over its regression and GAN baselines; the lead holds for real-world SR~\cite{stablesr_2024,seesr_2024} and remote sensing~\cite{ediffsr_2024}. Latent diffusion~\cite{rombach_2022} moved generation into a compressed latent space and spawned latent-space SR models: StableSR~\cite{stablesr_2024} adapts the pretrained Stable Diffusion prior with a time-aware encoder, ResShift~\cite{resshift_2023} trains a short chain that shifts residuals between LR and HR, and one-step distillations~\cite{sinsr_2024,osediff_2024} and semantics-guided variants~\cite{seesr_2024} push efficiency and fidelity further. Diffusion SR has also reached remote sensing~\cite{ediffsr_2024}. All of these models, whatever their backbone or sampler, share the fixed-crop constraint that motivates this paper: they generate patches, not scenes.

\paragraph{Large-content generation with diffusion models.}
To synthesize images beyond the training resolution, MultiDiffusion~\cite{multidiffusion_2023} fuses overlapping crop predictions by averaging at every step, and Mixture of Diffusers~\cite{mixture_2023} uses smooth spatial weights. SyncDiffusion~\cite{syncdiffusion_2023} aligns global style with gradient guidance. ScaleCrafter~\cite{scalecrafter_2024}, DemoFusion~\cite{demofusion_2024}, and AccDiffusion~\cite{accdiffusion_2024} target higher-resolution generation from pretrained latent models. These methods assume deterministic (ODE/DDIM-type) samplers, where averaging fuses only predicted signal. With stochastic SDE-type samplers, averaging also fuses independent noise. Sun~\etal~\cite{gvcf_2025} identified the resulting variance deficit in text-to-image generation and proposed guided, variance-corrected fusion (VCF). We build on this line in the \emph{conditional} SR setting and address a dimension none of these works consider: the memory and synchronization cost of joint denoising when the canvas is a gigapixel scene rather than a modest panorama.

\paragraph{Positioning.}
We contribute a fusion and inference layer rather than a new SR backbone. We instantiate it on SR3 deliberately. SR3 is the canonical pixel-space conditional DDPM with a stochastic ancestral sampler, so the fusion rule's effect can be isolated from confounds such as latent compression, text guidance, or backbone capacity. Our comparison (Sec.~\ref{sec:ablation}) varies \emph{only} the fusion rule under one checkpoint and seed. The variance-erosion analysis of Sec.~\ref{sec:erosion} applies to any sampler injecting independent per-tile Gaussian noise, and the correction carries over to latent-space samplers in principle.

\section{Method}
\label{sec:method}

InfScene-SR super-resolves a scene of arbitrary size with a patch-level conditional diffusion model by (i) denoising a set of overlapping tiles jointly, (ii) fusing overlaps with an exact variance-preserving rule, and (iii) organizing the computation so that every tile's update is local, additive, and independent of all other tiles. Fig.~\ref{fig:pipeline} shows the pipeline.

\subsection{Preliminaries: conditional diffusion SR}
\label{sec:prelim}
We adopt SR3~\cite{sr3_2023} as the backbone. Let $x$ be the LR condition (bicubically upsampled to the target grid) and $y_0$ the HR image. A network $f_\theta(x, y_t, \gamma_t)$, conditioned on the cumulative noise schedule $\gamma_t = \prod_{s\le t} \alpha_s$, is trained to predict the noise injected by the standard forward process~\cite{ho_denoising_2020,sr3_2023}. At inference, starting from $y_T \sim \mathcal{N}(0, I)$, each reverse step forms the posterior mean (posterior variance $\sigma_t^2$)
\begin{equation}
    \mu_t = \tfrac{1}{\sqrt{\alpha_t}} \Big( y_t - \tfrac{1-\alpha_t}{\sqrt{1-\gamma_t}}\, f_\theta(x, y_t, \gamma_t) \Big)
    \label{eq:mean}
\end{equation}
and samples
\begin{equation}
    y_{t-1} = \mu_t + \sigma_t \epsilon_t, \qquad \epsilon_t \sim \mathcal{N}(0, I).
    \label{eq:sample}
\end{equation}
The stochastic term in Eq.~\ref{eq:sample} distinguishes this \emph{ancestral (SDE-type)} sampler from deterministic ODE samplers such as DDIM~\cite{ddim_2020}. This term is what breaks naive joint denoising: when overlapping tiles are averaged, their independent noise partially cancels and the fused state no longer carries the expected variance $\sigma_t^2$.

\begin{figure}[ht]
    \centering
    \includegraphics[width=\linewidth]{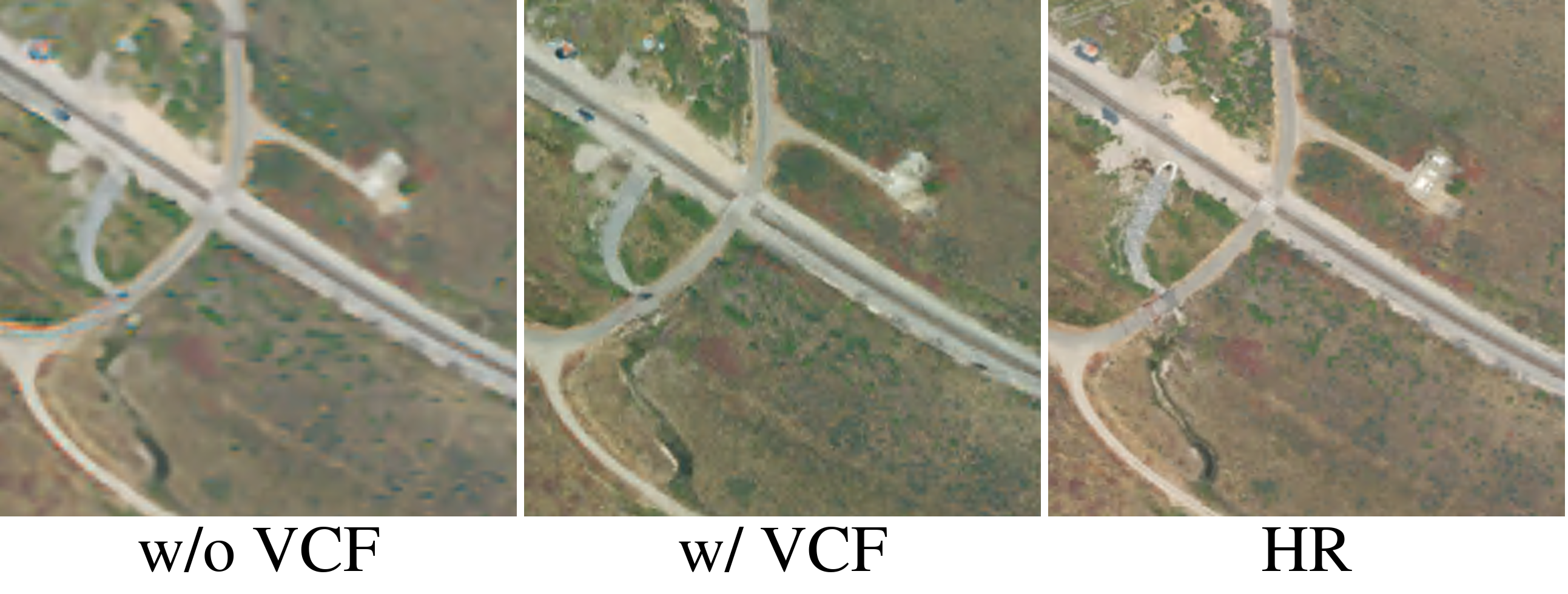}
    \caption{Variance erosion in the dense-overlap regime. Left: naive averaging (Eq.~\ref{eq:naive}) collapses stochastic detail into an over-smoothed reconstruction. Center: variance-corrected fusion (Eq.~\ref{eq:vcf}) restores sharp texture. Right: ground truth.}
    \label{fig:erosion}
\end{figure}

\subsection{Joint denoising and variance erosion}
\label{sec:erosion}
To super-resolve beyond the training crop, we cover the scene with overlapping tiles (views); $y_t^{(i)}$ and $x^{(i)}$ denote tile $i$'s state and condition. Joint denoising advances all tiles by one reverse step and reconciles overlaps before the next. Following GVCF~\cite{gvcf_2025}, each tile carries a spatial weight map $w_i$ peaking at its center. For a pixel covered by $N$ tiles, the naive fused state is
\begin{equation}
    \hat{y}_{t-1} = \frac{\sum_{i=1}^{N} w_i\, y_{t-1}^{(i)}}{\sum_{i=1}^{N} w_i}.
    \label{eq:naive}
\end{equation}

Each tile's sampled state $y_{t-1}^{(i)}$ carries its own noise $\sigma_t \epsilon_t^{(i)}$, independent across tiles. In overlaps the predicted means are strongly correlated ($\mu_t^{(i)} \approx \mu_t$), so the fused pixel has variance
\begin{equation}
    \operatorname{Var}(\hat{y}_{t-1})
    = \sigma_t^2\, \frac{\sum_i w_i^2}{\big(\sum_i w_i\big)^2}
    = \rho\, \sigma_t^2, \qquad \rho < 1 \text{ for } N > 1
    \label{eq:erosion}
\end{equation}
for nonnegative weights: the fused pixel carries less variance than the model was trained to invert, $\operatorname{Var}(\hat{y}_{t-1}) < \sigma_t^2$. We call this deficit \emph{variance erosion}. Compounded over $T$ steps, it destroys the detail SR should recover: dense overlap collapses generation toward the over-smoothed conditional mean (Fig.~\ref{fig:erosion}); at sparser overlap, texture corrupts in a coverage-dependent pattern (Sec.~\ref{sec:ablation}). Deterministic ODE samplers~\cite{ddim_2020} are unaffected, since Eq.~\ref{eq:naive} then averages only predicted means instead of variances.

\subsection{Variance-corrected fusion for conditional SR}
\label{sec:vcf}
Variance-corrected fusion (VCF)~\cite{gvcf_2025} repairs Eq.~\ref{eq:naive} by renormalizing the stochastic component so the fused pixel carries variance exactly $\sigma_t^2$ with unchanged expectation. With $W \equiv \sum_i w_i$ and $S \equiv \sqrt{\sum_i w_i^2}$,
\begin{equation}
    y_{t-1}
    = \frac{\sum_{i} w_i\, y_{t-1}^{(i)}}{S}
    + \Big(1 - \frac{W}{S}\Big)\, \frac{\sum_{i} w_i\, \mu_t^{(i)}}{W}.
    \label{eq:vcf}
\end{equation}
In the conditional setting each tile is anchored by its condition $x^{(i)}$, so overlapping means agree well and the correction applies cleanly. Where tiles do not overlap, $W = S = w_i$ and Eq.~\ref{eq:vcf} reduces to the ordinary sampler step.

\subsection{Spatially-decoupled variance correction}
\label{sec:sdvc}
Eq.~\ref{eq:vcf} is correct but scales poorly to whole scenes. Both normalizations are global, so a direct implementation gathers two full tensors per tile ($y_{t-1}^{(i)}$, $\mu_t^{(i)}$) into one memory space every step: tens of gigabytes of resident state per step for a gigapixel scene, behind a hard centralization point. This coupling is only apparent. The normalization maps
\begin{equation}
    W[p] = \sum_{j} w_j[p], \qquad S[p] = \Big(\sum_{j} w_j[p]^2\Big)^{1/2}
    \label{eq:maps}
\end{equation}
depend only on tiling geometry ($p$ indexes canvas pixels; the sums run over the tiles covering $p$), so they are precomputed once. Substituting them into Eq.~\ref{eq:vcf} and pushing the divisions inside the summation yields
\begin{equation}
    y_{t-1} = \sum_{i} C_t^{(i)}, \quad
    C_t^{(i)} = \tfrac{w_i}{S}\, y_{t-1}^{(i)} + \tfrac{S - W}{S\,W}\, w_i\, \mu_t^{(i)},
    \label{eq:sdvc}
\end{equation}
with $W, S$ restricted to tile $i$'s footprint. Every factor of $C_t^{(i)}$ is local to tile $i$ or precomputed, so the fused state is a plain sum. A contribution can be computed on any device and added at any time during the step, and no worker ever reads another tile's prediction. We call this form \emph{Spatially-Decoupled Variance Correction} (SDVC). The algebra is elementary once $W$ and $S$ are recognized as geometric constants, and the rewrite is exact: Eq.~\ref{eq:sdvc} produces the same sample as Eq.~\ref{eq:vcf}, but without a global stage, which makes the design of Sec.~\ref{sec:cost} possible.

\subsection{Cost model and distributed inference}
\label{sec:cost}
SDVC turns a reverse step into a map--reduce. \emph{Map}, per tile: read crops of the canvas, condition, and normalization maps; run the sampler step; form $C_t^{(i)}$. \emph{Reduce}: add $C_t^{(i)}$ into an accumulator canvas. This addition is the only computation outside the GPU, and it commutes, so contributions can arrive from any worker in any order. Once every tile's contribution is in, the accumulator is the complete $y_{t-1}$ and becomes the canvas for the next step.

The canvas, accumulator, and normalization maps are $\mathcal{O}(HW)$ but are touched only by crop reads and additive writes, so they live in host memory or on disk, and a stopped run resumes from them. The GPU holds one batch of tiles; its memory is set by the backbone and the batch size, not by the scene. Workers on any number of GPUs or nodes take disjoint tiles and post contributions asynchronously.

\section{Experimental Setup}
\label{sec:experiments}

\subsection{Task and datasets}
\label{sec:datasets}
\begin{figure}[t]
    \centering
    \begin{minipage}{0.48\linewidth}
        \centering
        \includegraphics[width=\linewidth]{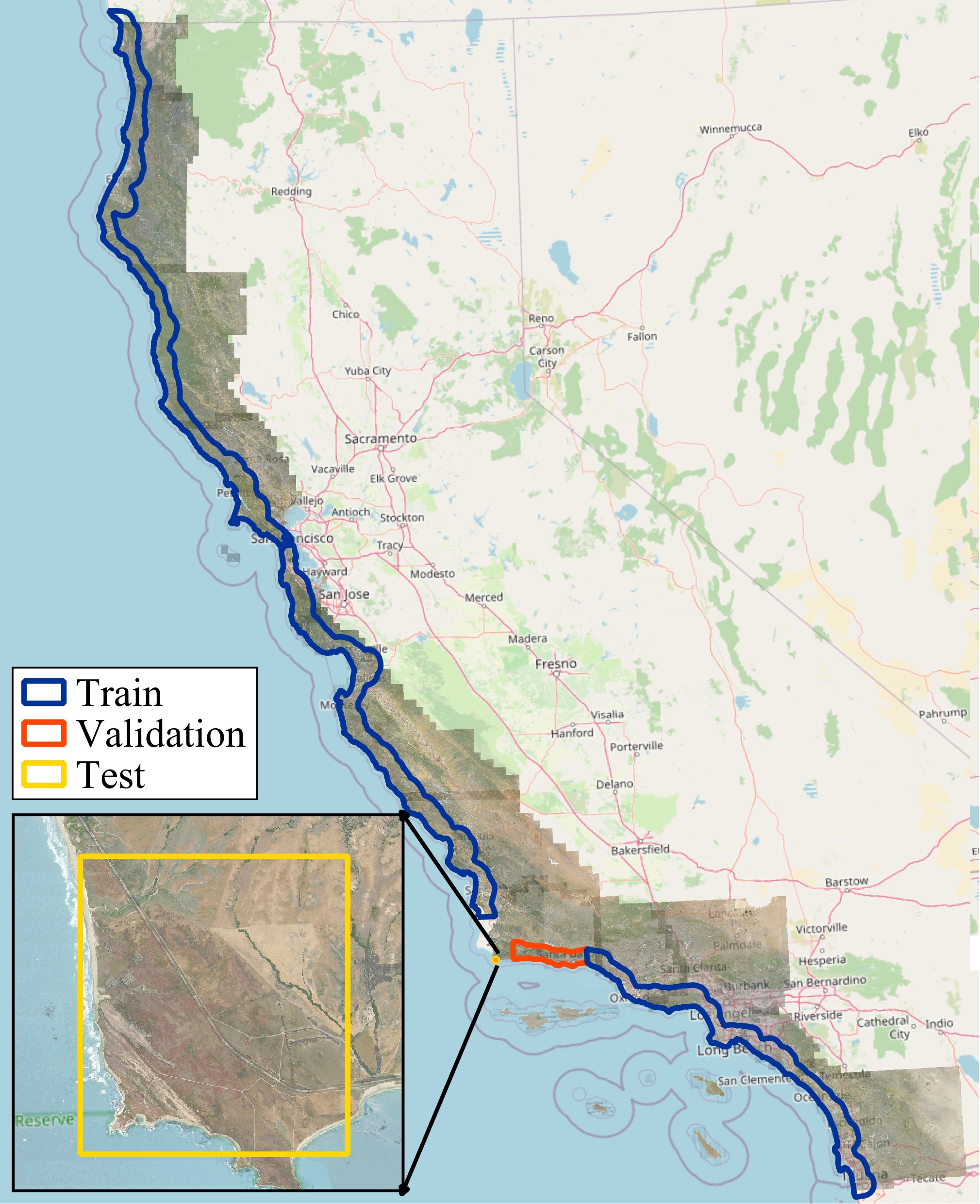}
    \end{minipage}\hfill
    \begin{minipage}{0.48\linewidth}
        \centering
        \includegraphics[width=\linewidth]{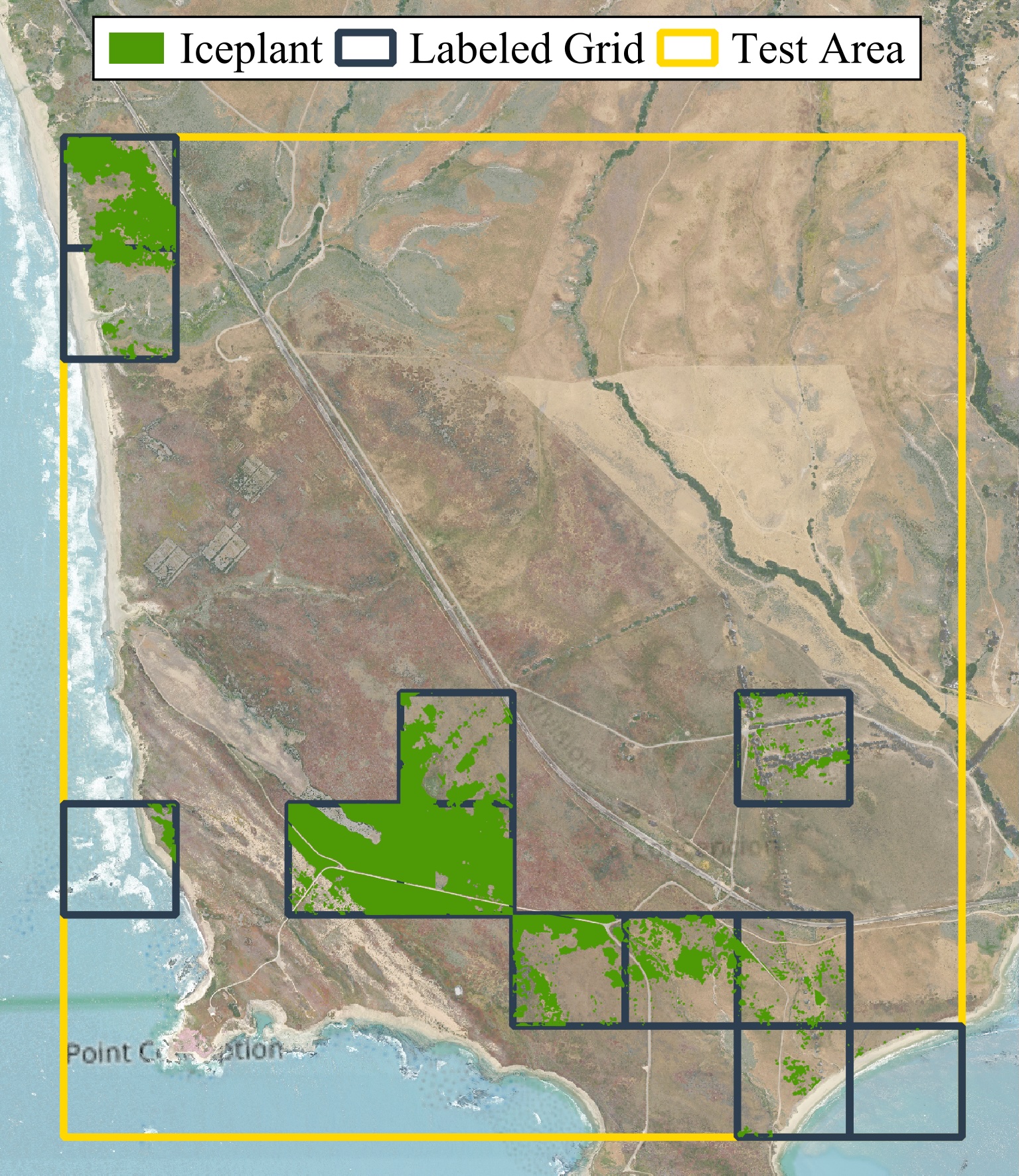}
    \end{minipage}
    \caption{Left: study area---2024 NAIP over 15 coastal California counties; Santa Barbara County held out. Right: expert iceplant annotations (green) on the 2024 test scene.}
    \label{fig:datasets}
\end{figure}

We build a 5$\times$ SR task from National Agriculture Imagery Program (NAIP) imagery~\cite{usda_naip}, mirroring the gap between 0.6\,m NAIP and ${\sim}3$\,m PlanetScope~\cite{planet_team_2017}. We downsample 0.6\,m NAIP to 3\,m (\texttt{INTER\_AREA}, then \texttt{INTER\_CUBIC} back to the HR grid) and train the model to invert this degradation. Training patches come from 2024 NAIP RGB imagery of 15 coastal California counties within a 10\,km coastal buffer (Fig.~\ref{fig:datasets}, left): 248{,}730 patches of $512{\times}512$ (236{,}633 train / 8{,}569 val / 3{,}528 test), with Santa Barbara County held out for validation and testing.

All whole-scene comparisons use a contiguous $4329{\times}4818$ test scene in Santa Barbara County (${\sim}21$ megapixels, ${\approx}80\times$ the training crop area).

\subsection{Whole-scene evaluation protocol}
\label{sec:protocol}
Every method first reconstructs the \emph{entire} scene, and all metrics are computed on that reconstruction. The order matters because seams are discontinuities between neighboring tiles: scored tile by tile, each tile looks fine on its own and the defect is never counted. We therefore compute each metric on the whole image whenever it allows; where a metric needs fixed-size inputs, we extract overlapping patches so that tile boundaries fall inside them rather than on their edges.

\textbf{Distortion and perceptual quality.} PSNR, SSIM~\cite{ssim_2004}, and LPIPS~\cite{lpips_2018} (AlexNet) are computed on the whole scene in one pass. FID~\cite{fid_2017}/KID~\cite{kid_2018} (\texttt{clean-fid}~\cite{cleanfid_2022}) are computed between $299^2$ patches extracted with 75\% overlap from reconstruction and reference; the dense extraction also stabilizes the estimates.

\textbf{Seam continuity.} Boundaries are a small fraction of the pixels, so they barely move a whole-image average. The \emph{Seam Gradient Index} (SGI) looks only at the interior tile-boundary lines $\mathcal{B}$ of the inference tiling (tile starts and ends, horizontal and vertical). For a line $b$, let $d(b)$ be the mean absolute difference between the pixels on its two sides, over its length and the three channels, and $\bar d(b)$ the median of the same quantity over the non-boundary lines within 24\,px on either side of $b$, skipping the 2\,px nearest to it. Then
\begin{equation}
    \mathrm{SGI} = \frac{1}{|\mathcal{B}|}\sum_{b\in\mathcal{B}} \frac{d(b)}{\bar d(b)}
    \;-\; \frac{1}{|\mathcal{B}|}\sum_{b\in\mathcal{B}} \frac{d_{\mathrm{HR}}(b)}{\bar d_{\mathrm{HR}}(b)},
    \label{eq:sgi}
\end{equation}
where the second term is the same average on the HR reference (1.01 on our scene), so native imagery scores 0. Positive values indicate seams, negative values over-smoothed boundaries.

\textbf{Observation consistency.} LR-PSNR is the PSNR between reconstruction and HR after both are projected to the LR grid (\cf\ the consistency score of SR3~\cite{sr3_2023} and LR-PSNR in SRFlow~\cite{srflow_2020}). LR-PSNR only checks whether the reconstruction still agrees with the LR observation. Detail finer than an LR pixel does not affect it. We report it next to PSNR because the pair locates the error: high LR-PSNR with low PSNR means the fine detail is placed differently from the reference, while low LR-PSNR means the reconstruction has drifted from the observation itself. A drop in LR-PSNR is therefore also the first sign of hallucinated structure at the scale of an LR pixel or larger (Sec.~\ref{sec:hallucination}).

\textbf{Downstream utility.} The downstream task is segmentation of the invasive iceplant (\textit{Carpobrotus edulis}), with expert annotations~\cite{iceplant_2024} as ground truth (Fig.~\ref{fig:datasets}, right). A U-Net~\cite{u-net_2015} with a ResNet-152~\cite{resnet_2016} encoder is trained on 4{,}913 patches ($128^2$) from 2022 NAIP imagery. It then segments each method's 2024 reconstruction with Gaussian-weighted sliding windows~\cite{resunet-a_2020} ($128^2$, stride 32), and Accuracy/Precision/Recall/F1/IoU are scored on the 261 annotated test patches of the scene.

\begin{table*}[t]
    \centering
    \footnotesize
    \begin{tabular}{l|c|cc|ccc|c|c}
        \toprule
        \textbf{Method} & \textbf{Stride} & \textbf{PSNR}$\uparrow$ & \textbf{SSIM}$\uparrow$ & \textbf{LPIPS}$\downarrow$ & \textbf{FID}$\downarrow$ & \textbf{KID}$\downarrow$ (${\times}10^{-2}$) & \textbf{LR-PSNR}$\uparrow$ & \textbf{SGI}\,$\to0$ \\
        \midrule
        Bicubic                      & --- & \textbf{28.77} & \textbf{0.715} & 0.569 & 90.13 & 7.20 & \textbf{42.03} & $\mathbf{-0.01}$ \\
        Independent + stitching      & 512 & 17.32 & 0.583 & 0.352 & 65.97 & 4.02 & 17.85 & 4.25 \\
        Independent + blending       & 384 & 17.66 & 0.617 & 0.342 & 65.95 & 4.07 & 18.14 & 0.88 \\
        Joint + naive averaging (Eq.~\ref{eq:naive}) & 384 & 21.50 & 0.480 & 0.649 & 121.30 & 9.78 & 25.30 & $-0.03$ \\
        \textbf{InfScene-SR (ours, Eq.~\ref{eq:sdvc})}  & 384 & 19.55 & 0.634 & \textbf{0.283} & \textbf{40.78} & \textbf{1.82} & 20.37 & 0.03 \\
        \textit{~~+ LF projection (Eq.~\ref{eq:lfproj})} & \textit{384} & \textit{27.07} & \textit{0.632} & \textit{0.217} & \textit{30.73} & \textit{0.95} & \textit{45.10} & \textit{0.03} \\
        \bottomrule
    \end{tabular}
    \caption{Whole-scene comparison on the $4329{\times}4818$ NAIP test scene (5$\times$ SR). All learned rows share one SR3 checkpoint and seed and differ only in the fusion strategy. Bold: best per column (closest to 0 for SGI, Sec.~\ref{sec:protocol}; bicubic is untiled and marks its zero point). The italic row post-processes our output using only the observed LR input (Sec.~\ref{sec:pd}); it is a diagnostic and is not ranked.}
    \label{tab:main}
\end{table*}

\subsection{Compared fusion strategies}
\label{sec:baselines}
All learned strategies share one SR3 backbone, checkpoint, seed, and tile size (512); all but stitching use stride 384 (25\% overlap). They differ only in how tiles interact. (1)~\textbf{Bicubic} is the upsampled LR input, seamless but blurry; it is also the SR condition. (2)~\textbf{Independent + stitching} is patch-based SR3~\cite{sr3_2023}: non-overlapping tiles (stride 512) are denoised in isolation and placed side by side. (3)~\textbf{Independent + blending}, the usual post-hoc fix, denoises overlapping tiles in isolation and composites them with the weights $w_i$ of Sec.~\ref{sec:erosion}. (4)~\textbf{Joint + naive averaging} fuses tiles at every step by Eq.~\ref{eq:naive}, as in ODE-based generation~\cite{multidiffusion_2023}. (5)~\textbf{InfScene-SR (ours)} fuses tiles at every step by Eq.~\ref{eq:sdvc}. Strategies 3--5 cost the same (one full reverse chain per tile at the same stride); stitching uses fewer tiles.

\subsection{Implementation details}
\label{sec:implementation}
The SR3 U-Net (base width 64, channel multipliers $[1,2,4,8,16]$) takes the condition and the noisy state concatenated along channels. Diffusion uses $T{=}2000$ steps with linear $\beta \in [10^{-6}, 10^{-2}]$. We train for 100 epochs with AdamW~\cite{adamw_2019} (lr $10^{-4}$, batch 8 per GPU, $8\times$ gradient accumulation, bf16). Whole-scene inference runs batch 8 per GPU under bf16 autocast, with the canvas, accumulator, and normalization maps kept on disk (Sec.~\ref{sec:cost}). Everything runs on one workstation with two L4 GPUs (24\,GB each). The downstream segmenter (Sec.~\ref{sec:protocol}) trains for 100 epochs with AdamW (lr $10^{-2}$ with exponential decay, $\gamma{=}0.97$; batch 64).

\section{Results}
\label{sec:results}

\subsection{Whole-scene comparison of fusion strategies}
\label{sec:ablation}
Table~\ref{tab:main} reports the whole-scene metrics of Sec.~\ref{sec:protocol} for all fusion strategies under the identical backbone. We make three observations.

\begin{figure*}[t]
    \centering
    \includegraphics[width=0.84\textwidth]{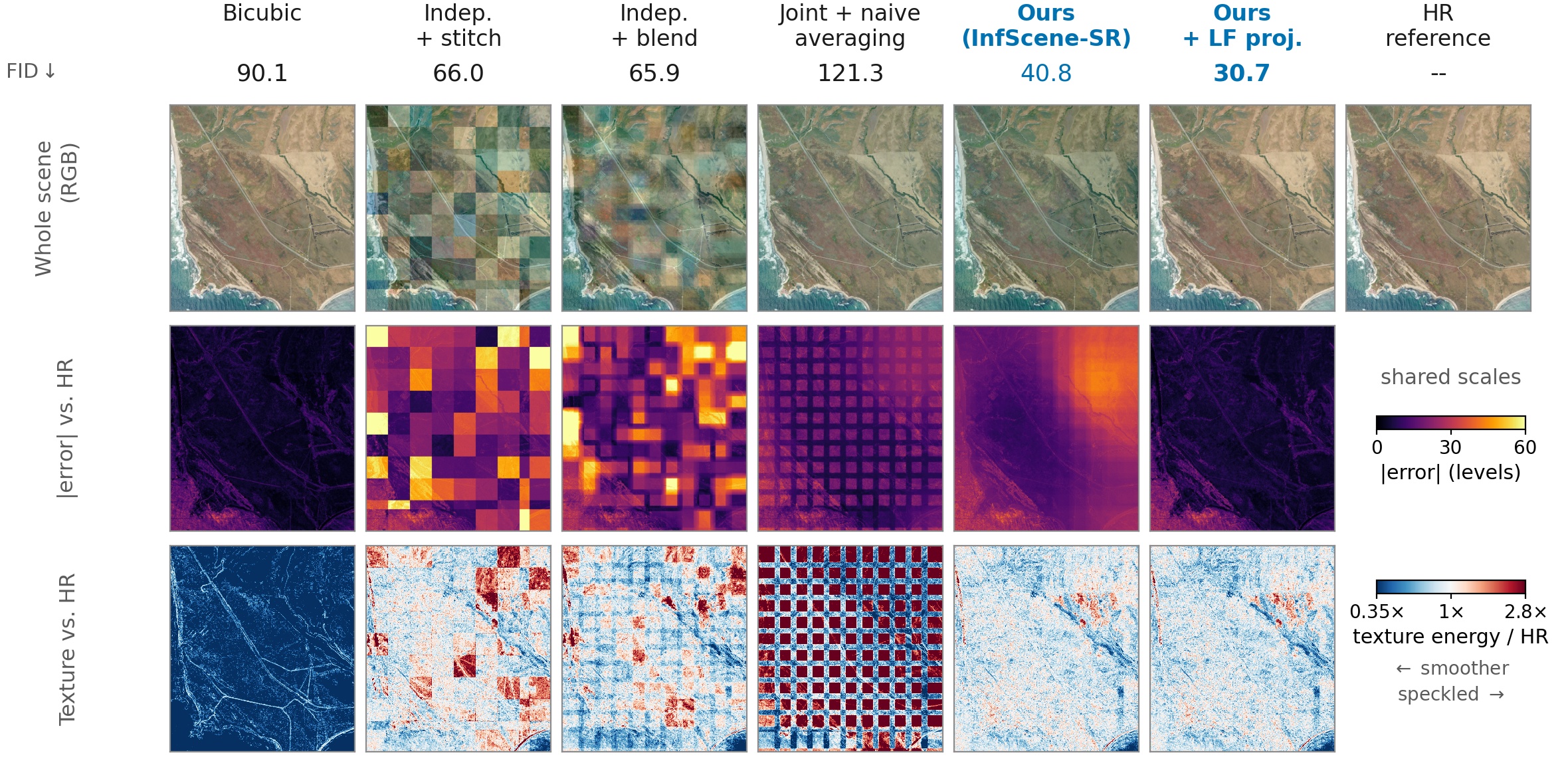}
    \caption{Whole-scene comparison under the identical backbone: reconstruction, per-pixel $|$error$|$, and local texture energy relative to HR.}
    \label{fig:results}
\end{figure*}

\textbf{Independent tiles fail everywhere, not just at boundaries.} Independent stitching has strong seams (SGI 4.25, i.e., boundary gradients five times those of native imagery), but seams are not its dominant error. Each tile is a separate sample of the conditional distribution, so the tiles drift apart radiometrically, and these whole-tile offsets depress fidelity across the scene. In the $|$error$|$ row of Fig.~\ref{fig:results}, the stitching column is a mosaic of whole-tile offsets. Feathered blending hides the hard edges but cannot reconcile content that diverged over the entire trajectory. Its boundaries still carry nearly twice the gradient of native imagery (SGI 0.88). Its FID of 66.0 equals stitching's, and since FID averages boundaries away, the equality shows that feathering changed the seams, not the content. Both independent strategies also drift furthest from the observation, with LR-PSNR at ${\sim}18$\,dB.

\textbf{Naive joint denoising trades seams for corrupted texture.} Averaging the tiles at every step keeps them from diverging, but it miscalibrates the sampler (Sec.~\ref{sec:erosion}). At dense overlap the whole output over-smooths (Fig.~\ref{fig:erosion}). At the 25\% overlap we use, the variance deficit varies with coverage and shows up in the texture row of Fig.~\ref{fig:results} as a waffle pattern, with over-smoothed overlap bands and speckled, mis-denoised tile interiors. The interiors are covered only once, yet they corrupt too. Naive averaging is seamless (SGI $-0.03$, boundaries slightly \emph{smoother} than native imagery) yet has the \emph{worst} perceptual quality in Table~\ref{tab:main}, FID 121.3 and LPIPS 0.649, below even bicubic. It buys coherence with the detail SR exists to provide, as Eq.~\ref{eq:erosion} predicts.

\textbf{Variance-corrected joint denoising achieves both.} InfScene-SR is the only strategy that is both seamless (SGI 0.03) and sharp (LPIPS 0.283 and FID 40.8, the best among raw reconstructions). In Fig.~\ref{fig:results}, our column shows no tile pattern in any row. The $|$error$|$ map varies only smoothly across the scene, and the texture map stays near $1\times$ throughout, with neither the blocks of the independent strategies nor the waffle of naive averaging. InfScene-SR also gains 1.9\,dB of PSNR over blending (19.5 vs.\ 17.7\,dB), because one mutually consistent sample drifts \emph{as a whole} rather than tile by tile. Coherent drift is recoverable, as we show next.

\subsection{Why bicubic wins PSNR: drift and texture price}
\label{sec:pd}
Every diffusion row in Table~\ref{tab:main}, ours included, scores below bicubic in PSNR and SSIM, yet bicubic trails ours on every perceptual metric. Stitching, which fuses nothing, shares the deficit, so it comes from the backbone's sampler rather than from the fusion layer. One part of it is the perception--distortion trade-off~\cite{blau_perception-distortion_2018}. SR3 itself trails its regression baseline in PSNR and SSIM on face SR yet wins human preference, because these metrics penalize any synthesized detail that is not perfectly aligned with the reference~\cite{sr3_2023}. A stochastic sampler draws one sample from the posterior and must \emph{place} high-frequency detail whose phase the LR image does not determine, so even correctly synthesized texture pays a per-pixel penalty. Bicubic adds nothing above the LR Nyquist frequency and pays none.

How much of our 9.2\,dB PSNR deficit to bicubic (19.5 vs.\ 28.8\,dB) is this price of texture? The LR-PSNR$-$PSNR gap (Sec.~\ref{sec:protocol}) answers this. For bicubic the gap is 13.3\,dB (28.8 vs.\ 42.0), so everything it lacks lies above the LR Nyquist frequency. For our raw output it is only 0.8\,dB (19.5 vs.\ 20.4), and for stitching 0.5\,dB. Our error survives area-averaging almost intact, so it is a low-frequency offset rather than misplaced texture. Ancestral sampling enforces no data consistency, and a whole-scene sample drifts radiometrically. Ours carries smooth, scene-coherent per-channel offsets, e.g., $-33$ gray levels in red. Unlike texture phase, everything below the LR Nyquist frequency \emph{is} observed, so this component can be re-imposed after sampling:
\begin{equation}
    y' \;=\; y \;+\; \mathrm{up}\big(\, x_{\mathrm{LR}} - \mathrm{down}(y) \,\big),
    \label{eq:lfproj}
\end{equation}
with $\mathrm{down}(\cdot)$ the area-average operator defining the LR grid, $\mathrm{up}(\cdot)$ bicubic, and $x_{\mathrm{LR}}$ the \emph{observed} LR image (no reference involved). We apply Eq.~\ref{eq:lfproj} once, after sampling, as a post-process that corrects the drift. It is a standard operation~\cite{backprojection_1991,dbpn_2018,ilvr_2021,ddrm_2022,ddnm_2023}, not a contribution, and we use it here as a diagnostic. It recovers 7.5\,dB of PSNR (19.5\,$\rightarrow$\,27.1) while \emph{improving} LPIPS (0.283\,$\rightarrow$\,0.217) and FID (40.8\,$\rightarrow$\,30.7). The LR-PSNR$-$PSNR gap opens to 18.0\,dB (27.1 vs.\ 45.1), so what remains is high-frequency. The 9.2\,dB deficit thus splits into ${\sim}7.5$\,dB of recoverable low-frequency drift and ${\sim}1.7$\,dB of texture-placement penalty, and the penalty buys $2.6\times$ better LPIPS and $2.9\times$ better FID than bicubic.

\subsection{Faithfulness and hallucination}
\label{sec:hallucination}
Generative SR raises a legitimate concern for scientific use: does the model invent structure? We check this in three ways. \textbf{Observation consistency.} Among the strategies that synthesize sharp texture, ours departs least from the observation: LR-PSNR 20.4\,dB against 18.1 for blending and 17.8 for stitching. The higher values in Table~\ref{tab:main} are trivial: bicubic (42.0\,dB) adds no detail, naive averaging (25.3\,dB) erases it, and the projection (45.1\,dB) enforces agreement by construction. Sec.~\ref{sec:pd} showed that our departure is a smooth scene-wide offset, not local disagreement. \textbf{Error geography.} In the $|$error$|$ row of Fig.~\ref{fig:results}, our raw column shows only that smooth offset; after the projection (next column) a faint, uniform grain of misplaced texture remains. An inserted or deleted object would show as a compact bright blob at object scale; neither column has one. The texture row adds that our local texture energy stays close to HR's ($1\times$), neither smoothed nor over-synthesized. \textbf{Downstream semantics.} A segmenter trained on HR imagery reaches IoU 0.755 on our reconstruction against 0.758 on HR itself (Table~\ref{tab:segmentation}), with higher precision than on the bicubic input (0.873 vs.\ 0.797). Hallucinated vegetation texture would create false iceplant and lower precision; it does not.

\subsection{Downstream iceplant segmentation}
\label{sec:downstream}
Table~\ref{tab:segmentation} reports iceplant segmentation on each input. IoU broadly follows the order of the perceptual metrics in Table~\ref{tab:main}: naive averaging 0.651 $<$ stitching 0.713 $<$ bicubic 0.725 $<$ blending 0.741 $<$ ours 0.755 $\approx$ HR 0.758. The strategies fail in different ways. Stitching's seams and per-tile drift fracture vegetation patches. Naive averaging is seamless, but its corrupted texture drops recall to 0.70, the worst of all. Both fall below the bicubic input. Bicubic fails the other way: it keeps HR's recall (0.889), but its blur makes the segmenter over-extend iceplant, so precision drops (0.797 vs.\ 0.837 on HR). InfScene-SR comes within 0.003 IoU of native HR, so we treat it as a tie. It trades recall (0.849 vs.\ 0.889) for precision (0.873 vs.\ 0.837), presumably because synthesized texture sharpens class boundaries. PSNR does not predict downstream utility, as SR3's authors found for ImageNet classification~\cite{sr3_2023}. Bicubic has the best PSNR but trails ours by 0.03 IoU; naive averaging beats our raw PSNR but segments worst; and the projection raises PSNR by 7.5\,dB (Sec.~\ref{sec:pd}) yet lowers IoU from 0.755 to 0.744. The projection changes radiometry, not texture, and the segmenter reads texture and edges. Which output suits a task therefore depends on what the task reads: the raw reconstruction for texture-driven analysis, the projected one where radiometry must agree with the observation, as in mosaicking and change detection.

\begin{table}[t]
    \centering
    \footnotesize
    \setlength{\tabcolsep}{4.5pt}%
    \begin{tabular}{l|ccccc}
        \toprule
        \textbf{Input} & \textbf{Acc.}$\uparrow$ & \textbf{Prec.}$\uparrow$ & \textbf{Rec.}$\uparrow$ & \textbf{F1}$\uparrow$ & \textbf{IoU}$\uparrow$ \\
        \midrule
        HR (native)          & 0.911 & 0.837 & 0.889 & 0.862 & 0.758 \\
        Bicubic              & 0.895 & 0.797 & \textbf{0.889} & 0.841 & 0.725 \\
        Indep.\ + stitching  & 0.896 & 0.837 & 0.828 & 0.833 & 0.713 \\
        Indep.\ + blending   & 0.906 & 0.838 & 0.864 & 0.851 & 0.741 \\
        Joint + naive avg.\  & 0.883 & \textbf{0.903} & 0.700 & 0.788 & 0.651 \\
        \textbf{InfScene-SR} & \textbf{0.914} & 0.873 & 0.849 & \textbf{0.861} & \textbf{0.755} \\
        \textit{~~+ LF projection} & \textit{0.909} & \textit{0.856} & \textit{0.849} & \textit{0.853} & \textit{0.744} \\
        \bottomrule
    \end{tabular}
    \caption{Downstream iceplant segmentation (HR-trained U-Net, 2024 test scene). Bold: best per column among reconstructions (HR is the reference, not ranked).}
    \label{tab:segmentation}
\end{table}

\subsection{Efficiency and scalability}
\label{sec:efficiency}
\begin{figure}[t]
    \centering
    \includegraphics[width=\linewidth]{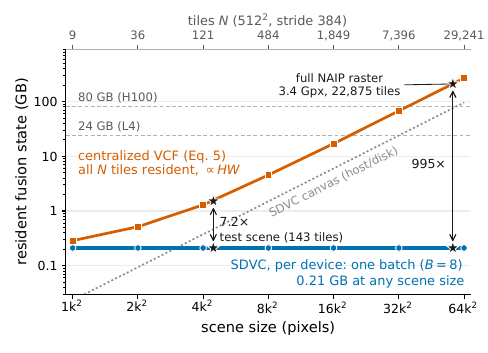}
    \caption{Resident GPU memory of the fusion layer per reverse step (float32; backbone excluded). Centralized VCF (Eq.~\ref{eq:vcf}) must hold every tile's state and mean, so its footprint grows with scene area; SDVC holds one batch of tiles at any size ($7\times$ less on the 143-tile test scene). Dotted: the $\mathcal{O}(HW)$ canvas and accumulator, streamed from host or disk; $\star$: the full 3.4-Gpx NAIP raster of the test county.}
    \label{fig:efficiency}
\end{figure}

Fig.~\ref{fig:efficiency} compares the GPU memory that the two forms of the corrected fusion, centralized VCF (Eq.~\ref{eq:vcf}) and SDVC (Eq.~\ref{eq:sdvc}), must keep resident per reverse step. The count covers the fusion state only, in float32; the backbone is excluded because both forms run the same sampler step on the same tiles, and the two forms are exact equivalents (Sec.~\ref{sec:sdvc}). \textbf{Memory.} Centralized VCF gathers every tile's state and mean before normalizing, so its GPU footprint grows with scene area. It needs 17\,GB at $16$k$^2$, exceeds the 24\,GB of an L4 at ${\sim}20$k$^2$, and would need 210\,GB for the full $70$k$\times48$k NAIP raster of our test county (3.4\,Gpx, 22{,}875 tiles). SDVC keeps one batch of tiles on each GPU whatever the scene size (nine tile-sized tensors per tile, 0.21\,GB for $B{=}8$), $995\times$ less at 3.4\,Gpx. The $\mathcal{O}(HW)$ canvas and accumulator (6\,GB at $16$k$^2$, 76\,GB for the full raster) live in host memory or on disk. The measured peak GPU memory of the whole pipeline, backbone included, stays at 13.1--13.3\,GB from $2$k$^2$ to $16$k$^2$. \textbf{Throughput.} Wall-clock time per step grows linearly with the number of tiles, at ${\sim}0.11$\,s per tile, measured from 36 to 1{,}849 tiles. The 143-tile test scene takes 16.3\,s per step, ${\sim}9$\,h at $T{=}2000$ on one L4, and a second GPU gives $1.8\times$ on the $4$k$^2$ benchmark (121 tiles, 13.1\,$\rightarrow$\,7.1\,s per step). Wall-clock is set by the backbone's 2000-step schedule, not by the fusion layer (Sec.~\ref{sec:limitations}).

\section{Limitations and Conclusion}
\label{sec:limitations}
\textbf{Limitations.} We test a single diffusion backbone. Adding a second, such as StableSR~\cite{stablesr_2024} or ResShift~\cite{resshift_2023}, means fine-tuning it on aerial imagery, which is beyond our compute budget of two L4 GPUs. Our inputs are NAIP downsampled to 3\,m rather than real PlanetScope imagery. A real PlanetScope--NAIP pair also differs in several respects, such as acquisition date and radiometric calibration, and those differences would enter the metrics alongside the fusion errors we set out to measure. Synthetic inputs hold everything but the fusion strategy fixed, and real sensor pairs are the natural next test. We also keep the backbone's 2000-step sampler as designed. SDVC removes joint denoising's memory and synchronization wall, not its sampling cost, so the 3.4-Gpx county of Sec.~\ref{sec:efficiency} would take about two months on one L4. Accelerated samplers are the complement we have not pursued.

\textbf{Conclusion.} InfScene-SR turns a fixed-crop diffusion SR model into a whole-scene super-resolver. Variance-corrected per-step fusion removes both seams and variance erosion, and the exact SDVC reformulation delivers constant per-device memory, resumable disk-streamed state, and asynchronous multi-GPU execution. Under one backbone it is the only fusion strategy that is seamless and sharp at once, the closest to the LR observation among those that synthesize detail, and within 0.003 IoU of native imagery on downstream segmentation.

{
    \small
    \bibliographystyle{ieeenat_fullname}
    \bibliography{main}
}

\end{document}


\maketitle
\setcounter{section}{0}
\renewcommand{\thesection}{\Alph{section}}
\renewcommand{\thetable}{S\arabic{table}}
\renewcommand{\thefigure}{S\arabic{figure}}

\section{Full derivation of SDVC}
\label{supp:derivation}
We restate the variance-corrected fusion (Eq.~5 of the main paper) for a pixel $p$ covered by tiles $i = 1,\dots,N$, writing $W = \sum_i w_i$ and $S = \sqrt{\sum_i w_i^2}$ (all quantities evaluated at $p$):
\begin{equation}
    y_{t-1}
    = \frac{\sum_{i} w_i\, y_{t-1}^{(i)}}{S}
    + \Big(1 - \frac{W}{S}\Big) \frac{\sum_{i} w_i\, \mu_t^{(i)}}{W}.
    \tag{S1}
    \label{eq:s1}
\end{equation}
Because $W$ and $S$ depend only on the tiling geometry, they are constants of the summation. Distributing the divisions,
\begin{align}
    y_{t-1}
    &= \sum_i \frac{w_i}{S}\, y_{t-1}^{(i)}
    + \frac{S - W}{S}\cdot\frac{1}{W} \sum_i w_i\, \mu_t^{(i)} \notag \\
    &= \sum_i \Big[ \underbrace{\frac{w_i}{S}\, y_{t-1}^{(i)}
    + \frac{S - W}{S\,W}\, w_i\, \mu_t^{(i)}}_{C_t^{(i)}} \Big],
    \tag{S2}
    \label{eq:s2}
\end{align}
which is Eq.~7 of the main paper. Every factor of $C_t^{(i)}$ is local to tile $i$ (its state, mean, and weight-map crop) or precomputable ($W, S$), so the fused state is a plain sum of independent contributions and the rewrite is exact.

\textbf{Mean and variance preservation.} Taking expectations of Eq.~\eqref{eq:s1} with $y_{t-1}^{(i)} = \mu_t^{(i)} + \sigma_t \epsilon_t^{(i)}$, $\epsilon_t^{(i)} \sim \mathcal{N}(0, I)$ i.i.d., and $\mu_t^{(i)} \approx \mu_t$ in overlaps,
$\mathbb{E}[y_{t-1}] = \frac{W}{S}\mu_t + (1 - \frac{W}{S})\mu_t = \mu_t$ and
$\operatorname{Var}(y_{t-1}) = \frac{\sigma_t^2 \sum_i w_i^2}{S^2} = \sigma_t^2$,
i.e., the corrected fusion restores exactly the marginal the sampler expects, whereas naive averaging yields $\rho\,\sigma_t^2$ with $\rho = \sum_i w_i^2 / W^2 < 1$ (main paper Eq.~4).

\section{Extended quantitative results}
\label{supp:tables}

Table~\ref{tab:supp_main} extends main paper Table~1 with the LF-projected variant (main paper Eq.~9) of every fusion strategy, not only ours. The projection uses the observed LR image alone, so it applies to any reconstruction, and it raises the PSNR of all four. It cannot repair the tiling. A seam is a high-frequency discontinuity and the projection only rewrites the low-frequency band, so stitching still scores SGI 3.36 and blending 0.70, against 0.03 for ours. Naive averaging still scores LPIPS 0.635 and FID 113.9, because eroded texture is not recoverable from the observation either. Blending edges ours in projected PSNR, 27.39 against 27.07, while ours holds every perceptual metric and the lowest SGI.

\begin{table*}[t]
    \centering
    \footnotesize
    \begin{tabular}{l|c|cc|ccc|c|c}
        \toprule
        \textbf{Method} & \textbf{Stride} & \textbf{PSNR}$\uparrow$ & \textbf{SSIM}$\uparrow$ & \textbf{LPIPS}$\downarrow$ & \textbf{FID}$\downarrow$ & \textbf{KID}$\downarrow$ (${\times}10^{-2}$) & \textbf{LR-PSNR}$\uparrow$ & \textbf{SGI}\,$\to0$ \\
        \midrule
        Bicubic                      & --- & \textbf{28.77} & \textbf{0.715} & 0.569 & 90.13 & 7.20 & \textbf{42.03} & $-0.01$ \\
        Independent + stitching      & 512 & 17.32 & 0.583 & 0.352 & 65.97 & 4.02 & 17.85 & 4.25 \\
        Independent + blending       & 384 & 17.66 & 0.617 & 0.342 & 65.95 & 4.07 & 18.14 & 0.88 \\
        Joint + naive averaging      & 384 & 21.50 & 0.480 & 0.649 & 121.30 & 9.78 & 25.30 & $-0.03$ \\
        \textbf{InfScene-SR}         & 384 & 19.55 & 0.634 & \textbf{0.283} & \textbf{40.78} & \textbf{1.82} & 20.37 & 0.03 \\
        \midrule
        \multicolumn{9}{l}{\emph{with LF projection (main paper Eq.~9):}} \\
        Independent + stitching      & 512 & 26.73 & 0.592 & 0.261 & 54.20 & 2.66 & 44.56 & 3.36 \\
        Independent + blending       & 384 & \textbf{27.39} & 0.625 & 0.254 & 43.92 & 2.27 & 44.88 & 0.70 \\
        Joint + naive averaging      & 384 & 23.79 & 0.470 & 0.635 & 113.87 & 8.86 & 44.87 & $-0.03$ \\
        \textbf{InfScene-SR}         & 384 & 27.07 & \textbf{0.632} & \textbf{0.217} & \textbf{30.73} & \textbf{0.95} & \textbf{45.10} & 0.03 \\
        \bottomrule
    \end{tabular}
    \caption{Complete whole-scene metric suite on the $4329{\times}4818$ NAIP test scene. The upper block is main paper Table~1 without its projected row; the lower block applies the LF projection to every strategy. Bold marks the best per column within each block. SGI is a diagnostic and is not ranked (target $0$; negative values mean boundaries \emph{smoother} than native imagery). After projection every strategy agrees with the observation by construction, so LR-PSNR no longer separates the rows.}
    \label{tab:supp_main}
\end{table*}

\begin{figure*}[t]
    \centering
    \includegraphics[width=1.00\textwidth]{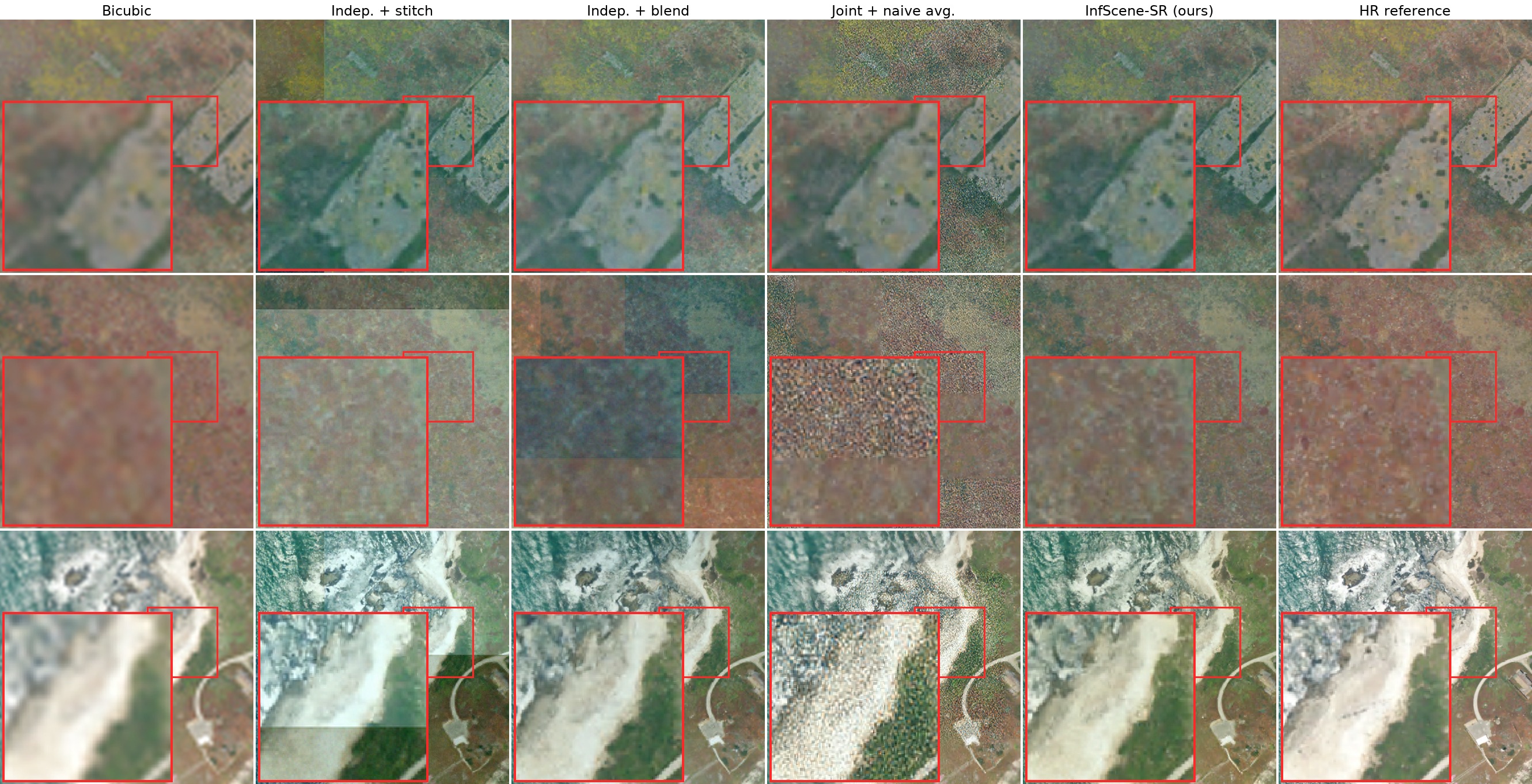}
    \caption{Crop-level comparison across three land-cover types. Rows: ruins and corrals, iceplant mosaic, surf zone. Each inset magnifies the marked square $2.4\times$.}
    \label{fig:supp_compare}
\end{figure*}

\begin{figure*}[t]
    \centering
    \includegraphics[width=\textwidth]{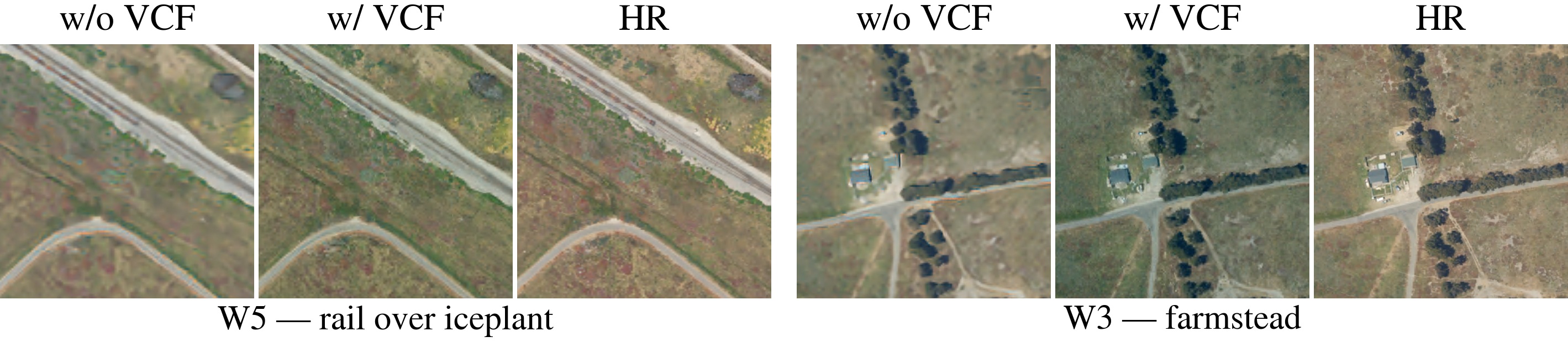}
    \caption{The dense-overlap collapse at one further location in each controlled window. Same controlled pairs as main paper Fig.~3, two $1024{\times}1024$ crops at patch $512$ and stride $128$ (up to $22$ tiles per pixel), one checkpoint and one seed ($42$), with \texttt{avg} against \texttt{sdvc} fusion as the only difference and no LF projection. W5 is the rail crossing, W3 the farmstead.}
    \label{fig:supp_erosion}
\end{figure*}

\section{Additional qualitative results}
\label{supp:qualitative}
Fig.~\ref{fig:supp_compare} compares the strategies at 1:1 scale on three land-cover types, which the whole-scene views of main paper Fig.~5 cannot resolve. Stitching lays a grid of tonal blocks over all three rows and a block edge crosses its surf inset. Blending softens those edges, but the blocks remain over the iceplant mosaic. Naive averaging replaces texture with a coarse speckle everywhere, worst in the surf row. InfScene-SR carries no block pattern and reads at the reference's sharpness. Fig.~\ref{fig:supp_erosion} repeats main paper Fig.~3 at one further location in each controlled window.